\documentclass[a4paper,twoside]{article}

\usepackage{epsfig}
\usepackage{subcaption}
\usepackage{calc}
\usepackage{amssymb}
\usepackage{amstext}
\usepackage{amsmath}
\usepackage{amsthm}
\usepackage{multicol}
\usepackage{pslatex}
\usepackage{apalike}
\usepackage{algorithm2e}
\usepackage[bottom]{footmisc}

\usepackage{multirow}
\usepackage{gensymb}
\usepackage{booktabs}
\usepackage{siunitx}
\usepackage{hyperref}
\usepackage{cleveref}

\usepackage{SCITEPRESS}     

\begin{document}

\title{MVTOP: Multi-View Transformer-based Object Pose-Estimation}

\author{\authorname{Lukas Ranftl\sup{1,2}\orcidAuthor{0009-0005-9248-1597}, Felix Brendel\sup{2}\orcidAuthor{0000-0002-1003-1877}, Bertram Drost\sup{2}\orcidAuthor{0000-0002-4109-6999} and Carsten Steger\sup{1,2}\orcidAuthor{0000-0003-3426-1703}}
\affiliation{\sup{1}School of Computation, Information and Technology, Technical University of Munich, Arcisstrasse 21, Munich, Germany}
\affiliation{\sup{2}MVTec Software GmbH, Munich, Germany}
\email{\{lukas.ranftl, felix.brendel, bertram.drost, carsten.steger\}@mvtec.com}
}


\abstract{We present MVTOP, a novel transformer-based method for multi-view rigid object pose estimation. Through an early fusion of the view-specific features, our method can resolve pose ambiguities that would be impossible to solve with a single view or with a post-processing of single-view poses. MVTOP models the multi-view geometry via lines of sight that emanate from the respective camera centers. While the method assumes the camera interior and relative orientations are known for a particular scene, they can vary for each inference. This makes the method versatile. Using lines of sight enables MVTOP to correctly predict the correct pose with the merged multi-view information. To show the model's capabilities, we provide a synthetic data set that can only be solved with such holistic multi-view approaches since the poses in the dataset cannot be solved with just one view. Our method outperforms single-view and all existing multi-view approaches on our dataset and achieves competitive results on the \hbox{YCB-V} dataset. To the best of our knowledge, no holistic multi-view method exists that can resolve such pose ambiguities reliably. Our model is end-to-end trainable and does not require any additional data, e.g., depth. The dataset will be publicly available at (https://www.mvtec.com/company/research/datasets/mvtec-mv-ball).}

\onecolumn \maketitle \normalsize \setcounter{footnote}{0} \vfill

\section{INTRODUCTION}
\label{sec:Introduction}
The six-degree-of-freedom (6-DoF) pose estimation task involves determining both the position $(x, y, z)$ and the orientation (represented, e.g., by pitch, yaw, and roll) of objects. It is an extension of object detection, where only a bounding box with center coordinates $(x, y)$ and size $(\textrm{width}, \textrm{height})$ is predicted. Estimation of the 6-DoF pose of objects remains crucial in tasks such as robotic manipulation \cite{busamStereoVisionApproach2015,wenCaTGraspLearningCategoryLevel2022,ghazaeiDealingAmbiguityRobotic2018,tremblayDeepObjectPose2018,kapplerRealTimePerceptionMeets2018,zhuSingleImage3D2014,colletMOPEDFrameworkObject2011}, augmented reality \cite{tang3DMapping6D2020,rambach6DoFObjectTracking2018,marchandPoseEstimationAugmented2016}, or industrial automation \cite{wu6DVNetEndToEnd6DoF2019,perezRobotGuidanceUsing2016}.

\begin{figure}[!t] 
    \centering
    \begin{minipage}[b]{0.49\linewidth}
        \centering
        \includegraphics[width=0.98\linewidth]{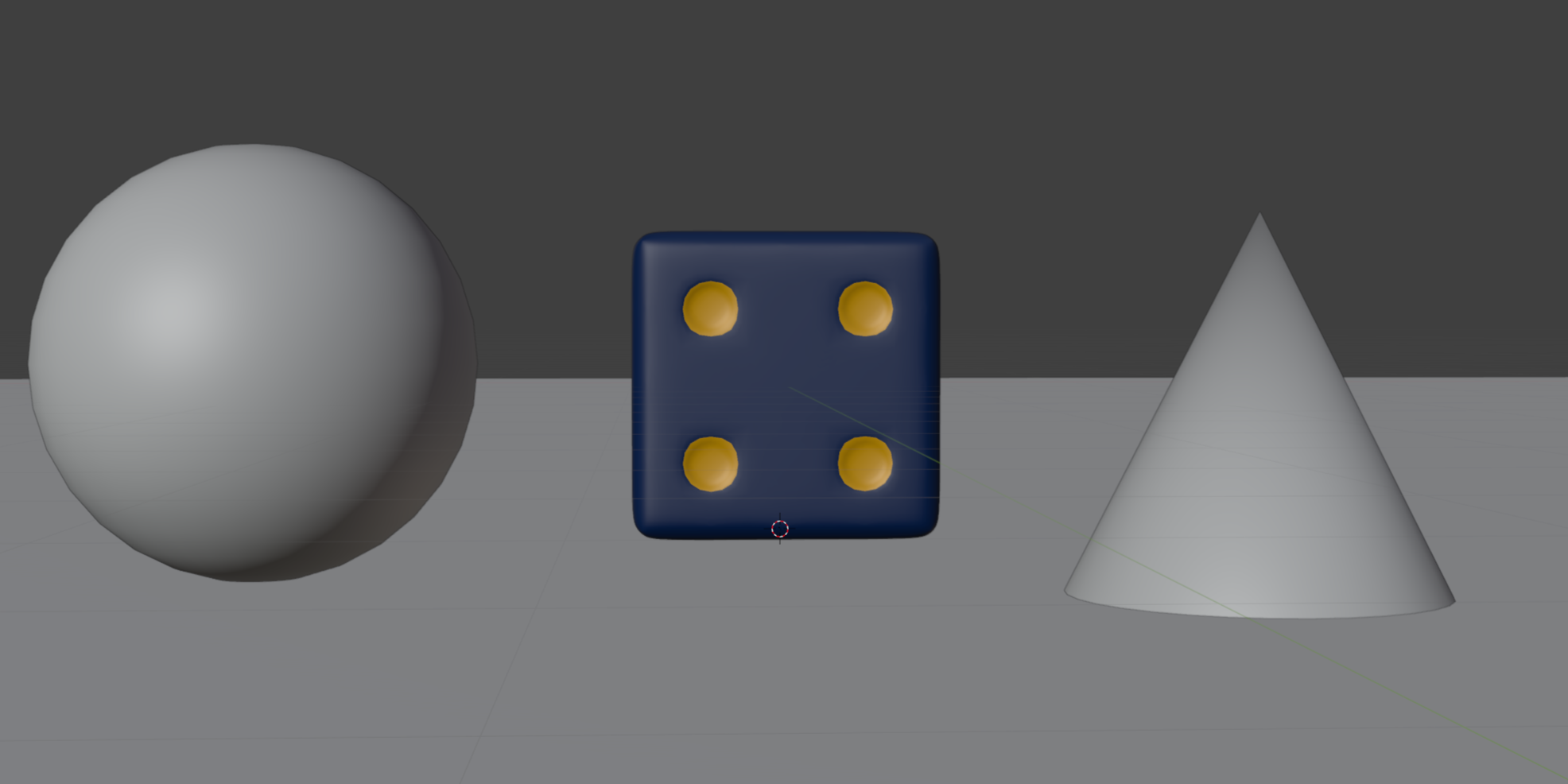}
        \caption*{a) View 1 of the die}
    \end{minipage}
    \hfill
    \begin{minipage}[b]{0.49\linewidth}
        \centering
        \includegraphics[width=0.98\linewidth]{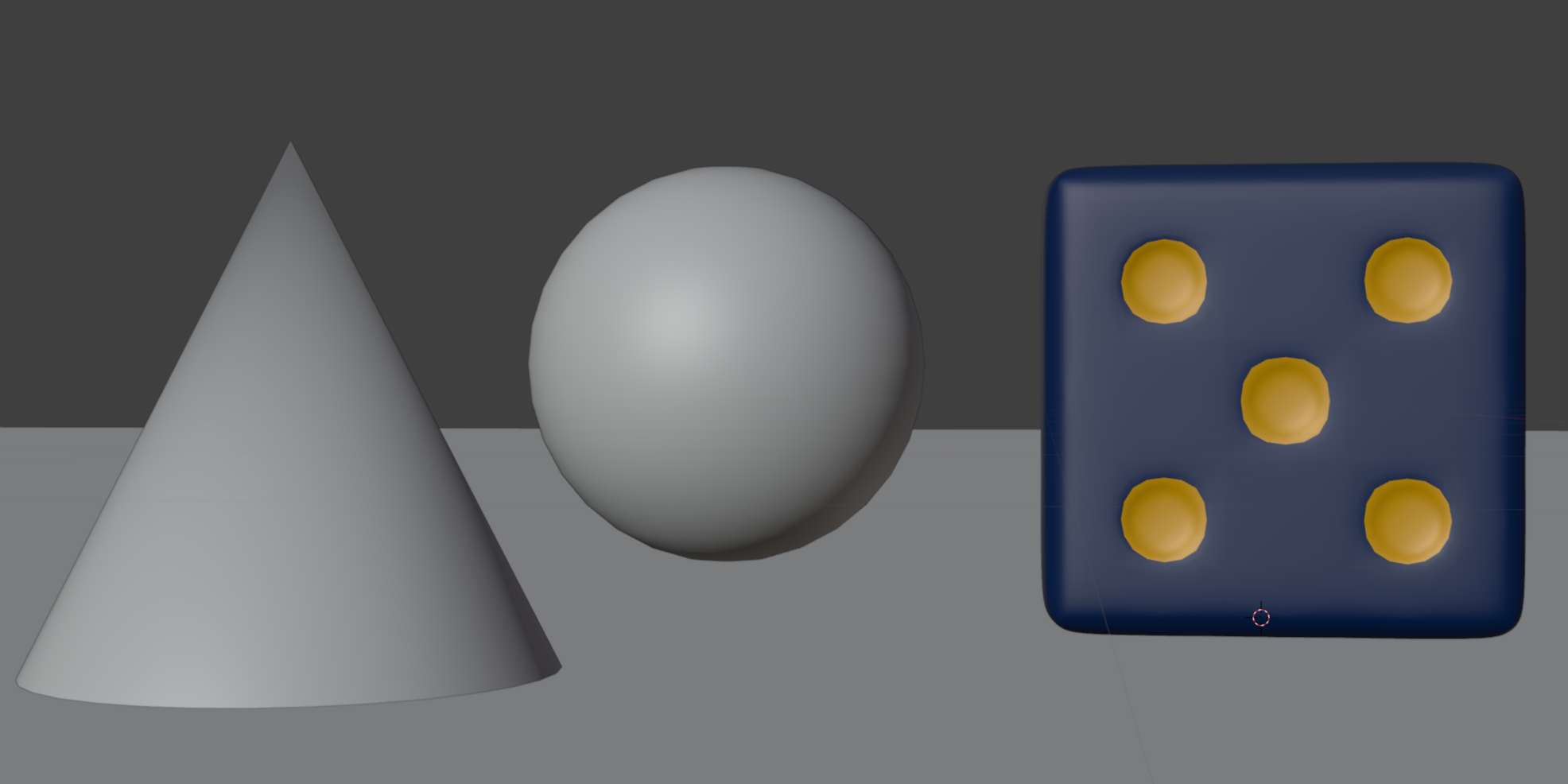}
        \caption*{b) View 2 of the die}
    \end{minipage}

    \vspace{1em} 

    \begin{minipage}[b]{0.98\linewidth}
        \centering
        \includegraphics[width=1\linewidth]{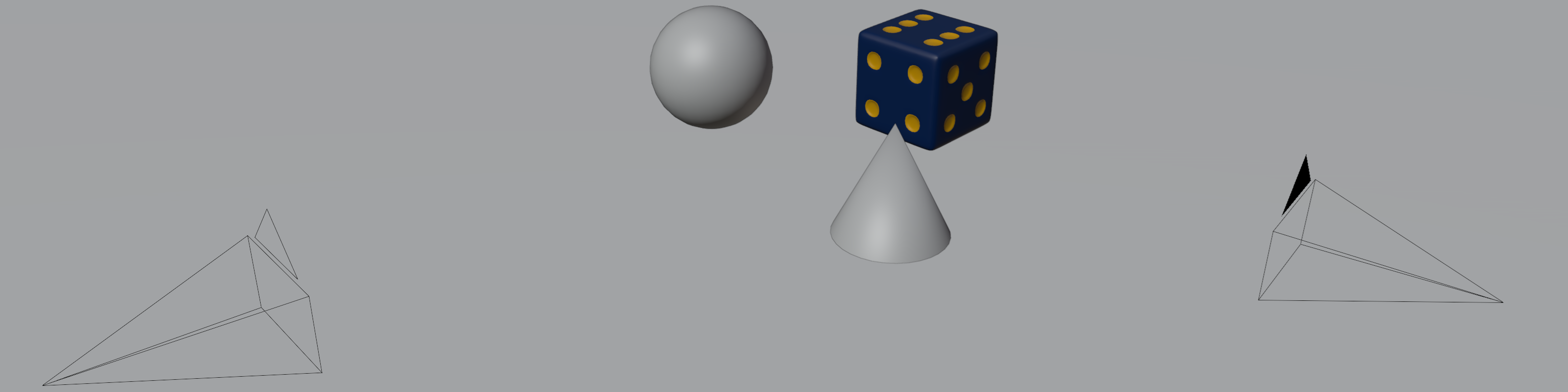}
        \caption*{c) View that shows the pose of the cameras and the die}
    \end{minipage}

    \caption{A die viewed from camera 1 (a), camera 2 (b), and from a larger distance with both cameras visible (c) (the cone and sphere are in the scene merely to aid in understanding the scene.) With the information of one view alone, the die's pose cannot be determined. It can only be determined if the information of both views is considered.}
    \label{fig:die}
\end{figure}

There are methods that approach this problem from a rule-based perspective, but the leaderboards of common benchmarks have been topped by deep-learning-based methods for some time now \cite{hodanBOPChallenge20202020,nguyenBOPChallenge20242025}. Usually, a backbone is used to extract image features. In indirect approaches, the features are used to generate 2D--3D correspondences that are used by the PnP algorithm to determine the object pose. In contrast, direct approaches use the features to directly regress the pose. Many methods pass additional depth information as an additional channel (RGB-D) to improve the pose estimate along the $z$-axis. However, using additional depth maps is related to higher financial costs since depth cameras are usually more expensive than RGB cameras, and it can also introduce a higher computational burden. Instead, multiple RGB images of the same scene can be used to determine the object pose. Because multiple views introduce geometric consistency, the depth estimate can be improved without requiring a depth map. Furthermore, pose ambiguities where the pose of an object cannot be inferred from one view can be solved with the help of additional views, e.g., a cup where the handle is not visible in the first view.
In situations where the pose cannot be inferred from one view alone, an early fusion and processing of the view-specific features is essential. This can be visualized using a regular die (\Cref{fig:die}). For each view of the die (Views~1 and~2), four possible poses are valid, i.e., four rotations around 90° along the current view axis. A correct pose estimation with either of these views is impossible. The correct die pose can be determined only by combining the information of both views, i.e., by knowing their relative camera poses. For the die, this pose ambiguity is discrete and limited to four possible poses. However, there are cases where the pose ambiguity is continuous (see \cref{sec:Dataset}). In the special case of the die, one can think of hand-crafted solutions where a multi-view pose estimation network could predict a set of poses (in this case four) for each view, and via a consistency check, the correct pose could be inferred; see, e.g., \cite{liUnifiedFrameworkMultiView2018}. However, such approaches may not generalize well and also fail for objects where the set of poses that match an observation is not discrete but continuous. Additionally, occlusion can induce ambiguities that cannot be enumerated exhaustively in an offline pre-processing stage. Overall, resolving pose ambiguities in a post-processing phase in which only poses from single views are considered is inherently very difficult.

Our presented network can tackle such ambiguities in an end-to-end manner without the need for fine-tuning it to the camera setup. Especially in an industrial context, the acquisition of depth data can be associated with excessive costs, as multiple RGB cameras are usually cheaper than one 3D-sensor, which is why our approach only requires RGB data. Furthermore, our multi-view approach can process the input views in an arbitrary order because the network learns to handle ray information generated by the internal and relative orientations of the cameras. The attention mechanism \cite{vaswaniAttentionAllYou2017} ensures an exchange of information between the different camera viewpoints and allows the handling of ambiguities as well as a better pose estimation in general.

Our main contributions are as follows:
\begin{itemize}
    \item We provide the novel synthetic multi-view pose estimation dataset named MV-ball that contains images of an object with pose ambiguities that are only solvable by early incorporation of information from multiple views. The dataset is used to demonstrate the advantages of our proposed method MVTOP.
    \item For the first time, we introduce a novel multi-view pose estimation framework that fuses view-specific features in an early stage and is end-to-end trainable. By merging line-of-sight information and multi-view features and processing them using the attention mechanism, the network can build a strong and discriminative scene understanding. This allows the model to determine the correct pose for objects that are unsolvable for single-view methods.
    \item Our method outperforms existing methods on the MV-ball dataset, which we introduce to highlight the above-mentioned ambiguities of poses. On the \hbox{YCB-V} dataset, we achieve competitive results.
    \item Our model uses only RGB images and the interior and relative orientations of the cameras. It does not require additional depth information or 3D models for Inference. 3D models are only required for the generation of training data. Furthermore, the views/images can be fed to the model in an arbitrary order since it learns to handle the relative orientations of the camera to build up a spatial understanding of the 3D scene.

\end{itemize}

\section{RELATED WORKS}
\label{sec:Related works}

\subsection{Object Pose Estimation}
The field of object pose estimation can be divided in various ways. Within the following overview we focus on the learning based single- and multi-view approaches that are fine-tuned on one or multiple objects. Here, there are distinctions in terms of model architecture, input data, and post-processing.

Most of the existing direct methods apply a CNN from which the poses are regressed directly \cite{kehlSSD6DMakingRGBBased2017,wangSelf6D16thEuropean2020,xiangPoseCNNConvolutionalNeural2018}. In \cite{sockIntroducingPoseConsistency2020}, the network uses a differentiable PnP head to find the object poses.

Many indirect methods infer the pose by applying the PnP algorithm on 2D--3D correspondences \cite{guptaCullNetCalibratedPose2019,zakharovDPOD6DPose2019,parkPix2PosePixelWiseCoordinate2019,hodanEPOSEstimating6D2020,wangGDRNetGeometryGuidedDirect2021,diSOPoseExploitingSelfOcclusion2021}. In \cite{sundermeyerImplicit3DOrientation2018}, an autoencoder is used together with an ICP module to predict the pose. In addition, there are approaches that use the 3D model to generate templates that are then matched with the objects in the image to find the poses \cite{nguyenTemplates3DObject2022,hinterstoisserGradientResponseMaps2012}. In \cite{suZebraPose2022IEEE2022}, the network predicts bit masks from which the pose is determined using the PnP algorithm.

Recently, attention-based transformer architectures have been explored for pose estimation. In \cite{aminiT6DDirectTransformersMultiobject2021}, the Detection Transformer DETR \cite{carionEndtoEndObjectDetection2020} is adapted to the task of estimating poses. Since this approach includes a symmetry-aware loss, the method needs 3D models of the objects. The method in \cite{jantosPoETPoseEstimation2023} is based on Deformable DETR \cite{zhuDeformableDETRDeformable2020} and requires only one RGB image to predict the pose. No 3D models or additional information are necessary and the network is end-to-end trainable.

The research in the field of multi-view pose estimation is rather sparse. The methods in \cite{zengMultiviewSelfsupervisedDeep2017,duffhaussMV6DMultiView6D2022,sockMultiview6DObject2017} are multi-view methods that require depth information. To our knowledge, so far there are only two methods that rely solely on RGB images \cite{labbeCosyPoseConsistentMultiview2020,liUnifiedFrameworkMultiView2018}. The network in \cite{labbeCosyPoseConsistentMultiview2020} consists of several stages. The first stage generates pose candidates for each view separately. For this, any single-view pose estimator could be used. In the next stage, the candidates are matched across the views using a RANSAC-based method. The matches are used to build a graph that links consistent poses. In a final step, the poses are refined by minimizing a multi-view reprojection error. However, this approach cannot deal with pose ambiguities as described in \cref{sec:Introduction} since the network does not fuse multi-view information before the pose prediction. In \cite{liUnifiedFrameworkMultiView2018}, the network produces a set of the top-K poses for each view. In a common reference frame, the pose hypotheses are compared and scored via a voting scheme. Therefore, ambiguities can be resolved using cross-view consistency, but only in discrete cases.

\subsection{Object Pose Estimation Datasets}
Several datasets have been proposed for object pose estimation. Many of them are part of the well-known BOP Challenge \cite{hodanBOPChallenge20202020,nguyenBOPChallenge20242025}. The most important ones are the core datasets of the main track, which have been part of the Challenge since 2019. The datasets within this track are Linemod-Occluded \cite{brachmannLearning6DObject2014}, T-LESS \cite{hodanTLESSRGBDDataset2017}, ITODD \cite{drostIntroducingMVTecITODD2017}, HomebrewedDB \cite{kaskmanHomebrewedDBRGBDDataset2019}, YCB-V \cite{xiangPoseCNNConvolutionalNeural2018}, IC-BIN \cite{doumanoglouRecovering6DObject2016}, and TUD Light \cite{hodanBOPBenchmark6D2018}. The test sets contain real RGB/monochrome (+depth) images and the train sets contain real and/or synthetic RGB/monochrome (+depth) images. The BOP-H3 track features the three datasets HOPE \cite{tyree6DoFPoseEstimation2022}, HANDAL \cite{guoHANDALDatasetRealWorld2023} and HOT3D \cite{banerjeeHOT3DHandObject2025}. These datasets focus on household objects and some include additional hand poses. Recently, an additional track (BOP-Industrial) was opened that focuses on industrial objects and also includes multiple views \cite{nguyenBOPChallenge20242025}. The datasets XYZ-IBD \cite{huangXYZIBDHighprecisionBinpicking2025}, IPD \cite{kalraCoEvaluationCamerasHDR2024}, and ITODD \cite{drostIntroducingMVTecITODD2017} were included in this track. Furthermore, the ROBI dataset \cite{yangROBIMultiViewDataset2021} models a bin picking scenario, where each bin is filled with one of seven industrial objects. Although there are already many datasets, to the best of our knowledge, only HOT3D \cite{banerjeeHOT3DHandObject2025} contains an object (a Rubik's cube) that is ambiguous from multiple views. However, the images in this dataset also contain objects that are not ambiguous. This motivated us to generate a dataset that covers exactly this gap.

\section{MV-BALL DATASET}
\label{sec:Dataset}
To evaluate the ability of a method to combine visual information from multiple views, we designed a dataset where single images can only result in ambiguous poses. To predict a correct pose, multiple views would have to be fused by necessity and therefore all single-view methods, as well as methods that naively fuse poses of single-view results, are expected to fail on this dataset. Existing datasets cannot highlight the nature of such ambiguous poses. In YCB-V \cite{xiangPoseCNNConvolutionalNeural2018}, objects such as a mug are ambiguous if the handle is not visible. As soon as the handle is visible, the pose can be correctly determined. In theory, this makes such objects solvable with a single view. No existing dataset contains objects that are ambiguous from multiple views.

\begin{figure}[!t]
    \centering
    \begin{minipage}[b]{0.48\linewidth}
        \centering
        \includegraphics[width=\linewidth]{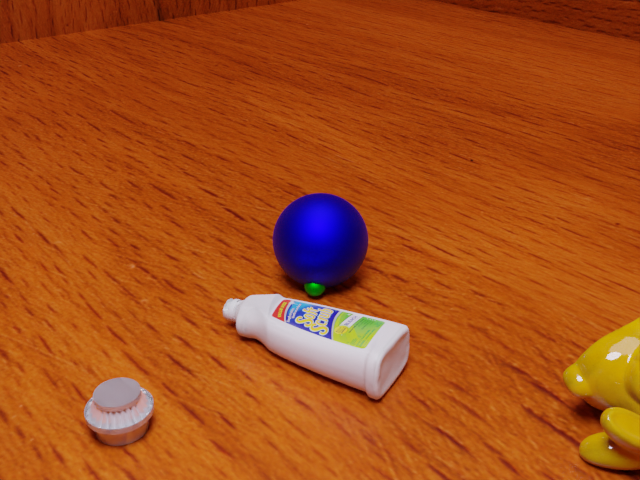}
        \subcaption{View 1 (only the green hemisphere is visible)} 
        \label{fig:green_example}
    \end{minipage}
    \hfill
    \begin{minipage}[b]{0.48\linewidth}
        \centering
        \includegraphics[width=\linewidth]{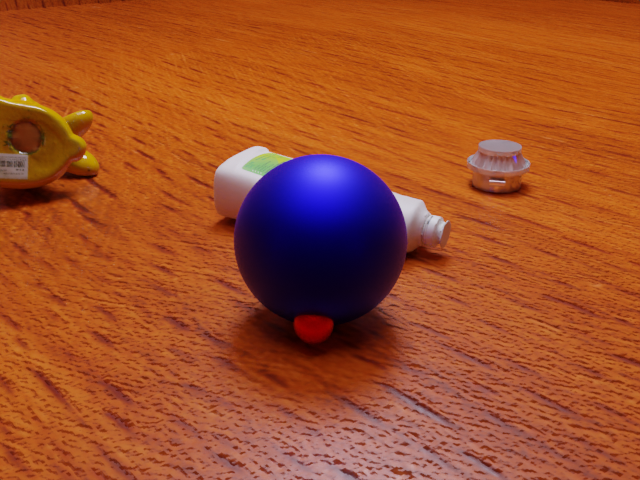}
        \subcaption{View 2 (only the red hemisphere is visible)} 
        \label{fig:red_example}
    \end{minipage}

    \caption{Example view pair of the MV-ball training split. View 2 differs by approximately 180° with respect to the axis pointing upwards from the floor}
    \label{fig:mvball_example}
\end{figure}

To allow for many ambiguous views, two hemispheres were extruded from a sphere at a relative angle of 90°, as can be seen in \cref{fig:mvball_example}. With this object, all views that do not include both hemispheres are ambiguous since the position of the second extruded hemisphere can only be guessed. To generate a dataset of ambiguous views, blenderproc \cite{denningerBlenderProcReducingReality2020} was used to place objects in scenes and let them fall to the floor together with random distractor objects. To avoid a bias in the dataset, the two hemispheres were modeled with no mass, which means the hemisphere would not rotate towards the ground due to the influence of gravity. This means that the two hemispheres do not always point towards the ground plane but also point upwards in the world, and a larger range of poses is covered. The resulting scenes were used to find camera poses in which only one of the two hemispheres is visible by setting pixel thresholds of how many pixels per hemisphere must be visible. In our setup, we used two thresholds to create test sets of different difficulty --- the easy test set with more than 300 pixels visibility and a hard test set with more than 10 pixels visibility for one hemisphere. In both cases the second hemisphere is not visible at all. The images have a resolution of 640$\times$480 pixels. Our dataset consists of 112.042 images for training and 3853 or 3458 images for testing/evaluating with the easy or hard set, respectively. With multiple instances in one image, some of them might be solvable with one view, while others would not be solvable at all (if no hemisphere is visible). Therefore, every image in our dataset contains only one instance to ensure that the metric for such an image would be distinct.  It can be noted that this dataset is only meant to showcase the true multi-view capabilities of pose estimation approaches while also dealing with self-occlusion

Although this is a purely synthetic dataset, it is more than well-suited to benchmark holistic multi-view approaches and to emphasize on the use case of such pose ambiguities. A real-world dataset covering this scenario has yet to be generated.

\begin{figure*}[t]  
  \centering
  \includegraphics[width=0.98\textwidth]{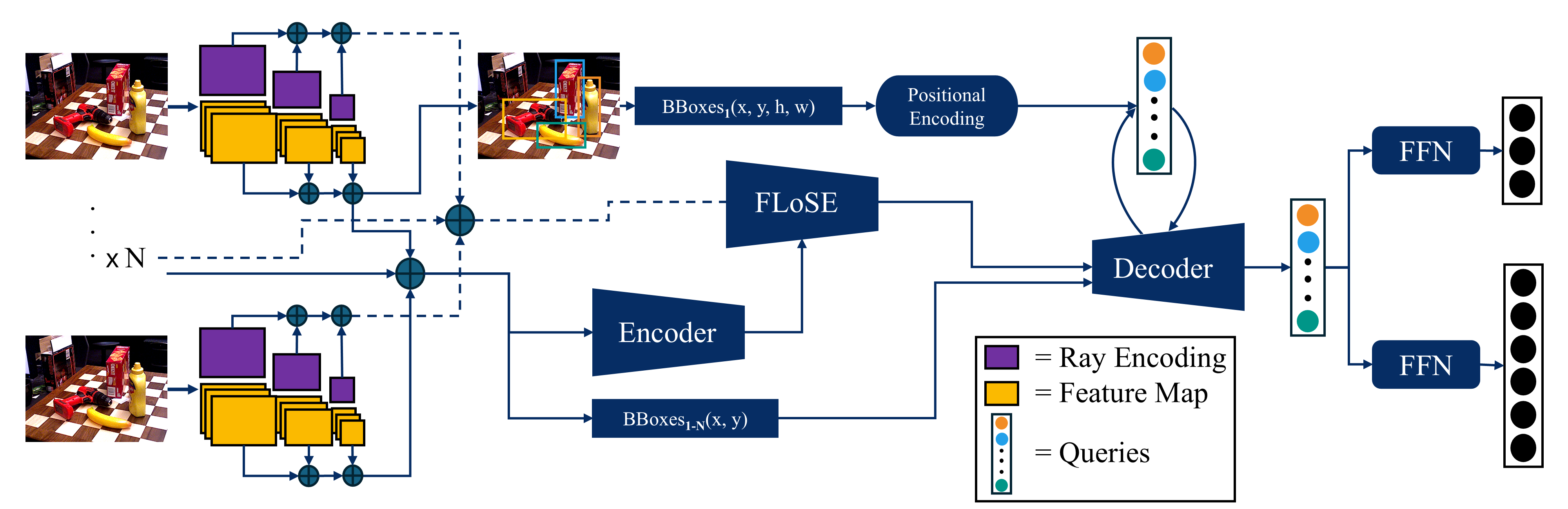}
  \caption{The MVTOP network. Our model takes N input images of different views. An object detector extracts multi-scale features (yellow) and predicts bounding box coordinates for each image. Additionally, the lines of sight for the feature maps are encoded (purple) using the interior and relative orientations of the cameras. By merging this line-of-sight information with the encoded multi-scale multi-view features, the decoder can produce a 3D aware update of the queries that were generated from the bounding box information of the first image. Finally, two separate heads predict rotation and translation for each query.}
  \label{fig:MVTOP_arc}
\end{figure*}

\section{MVTOP METHOD}
\label{sec:Method}
Our work is inspired by PoET \cite{jantosPoETPoseEstimation2023} and the human pose estimation network in \cite{wangDirectMultiviewMultiperson2021}. Both are based on Deformable-DETR \cite{zhuDeformableDETRDeformable2020}. In the predecessor DETR, features of the whole image are considered, whereas in Deformable-DETR, attention is established only between local features around specific reference points \cite{carionEndtoEndObjectDetection2020}. In \cite{wangDirectMultiviewMultiperson2021}, the poses of multiple people are predicted with information from multiple images. To improve generalization, the network uses so-called joint query embeddings, which encode person and joint information together. Furthermore, a module named projective attention is introduced, which takes local features around the 2D projections coming from 3D points. Thereby, the model fuses multi-view features. This is enriched with RayConv, which is a module that encodes camera ray directions into the feature vector.

Our network takes $N$ input images and processes them via an off-the-shelf object detector to extract multi-scale features and bounding box information. Additionally, the multi-scale features are enriched with line-of-sight information that encodes both the interior and relative orientations of the cameras. The gathered information is processed by an attention-based encoder--decoder transformer. The output of the transformer is processed with a rotation head and a translation head. Our method is visualized in \cref{fig:MVTOP_arc} and is described in detail in the following section. 

\subsection{Network Architecture}
First, the $N$ input images are passed to the object detector (OD). It processes each view separately and extracts multi-scale features, classes, and bounding-box information. The multi-scale features are fed into the attention-based encoder, which generates a suitable embedding for the decoder.

For every pixel in every feature map, the line of sight is calculated using the interior and relative orientations of the cameras. This information is merged with the encoded feature maps in a step which was introduced as RayConv in \cite{wangDirectMultiviewMultiperson2021}. We adapted this operation and call it FLoSE (Feature Line-of-Sight Encoding) which is explained in more detail below. The spatially enriched and encoded feature maps are processed by the decoder.

In the Decoder, the first image serves as a reference image for which the pose prediction will be made. The center values of the bounding boxes of the first image are used to generate a set of object queries, which will be processed by the decoder. The center values of each bounding box of each object in each image are given as reference points to the decoder. These reference points are used within the projective attention module to sample features in the region around the found objects. By not extracting dense features but only features within the vicinity of the objects of interest, computational cost can be reduced. Furthermore, the gathered features are more discriminative.

Finally, the rotation and translation heads predict the pose components for each query for the first image. The translation and rotation heads are unchanged compared to the implementation of PoET \cite{jantosPoETPoseEstimation2023}.

\begin{figure}[!t] 
    \begin{minipage}[b]{0.98\linewidth}
        \centering
        \includegraphics[width=1\linewidth]{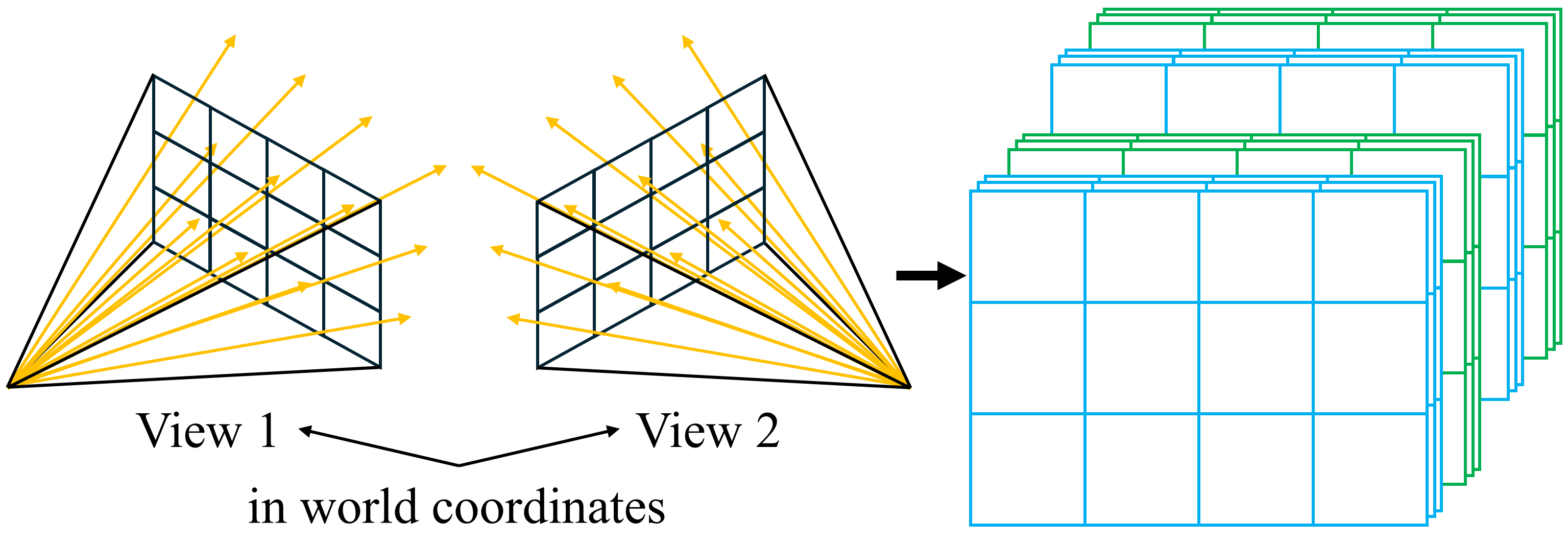}
    \end{minipage}
    \caption{The lines-of-sight (LoS - here in yellow) are generated for each pixel (here 3x4 example) in every views using the intrinsic and extrinsic camera parameters. Every LoS can be encoded using 6 values; 3 for the origin (blue) and 3 for the direction (green) of the ray.} 
    \label{fig:LoS_visualization}
\end{figure}

The center values of the objects in each view allow only limited 3D scene understanding. To enrich the model with more spatial information, we adapt the RayConv operation of \cite{wangDirectMultiviewMultiperson2021} (see also \cref{fig:LoS_visualization}). Our goal was to train a model that is capable of dealing with arbitrary camera setups and sequences of camera input. To do so, the lines of sight are calculated for the feature maps using the interior and relative orientations of the cameras. One possible parameterization consists of the origin and direction. This results in six values for each pixel. Other parameterizations are examined in \cref{sec:Ablations}. By randomly combining views in each iteration, the model learns to handle a new camera setup at inference. 
For every pixel in every feature map $ l $ in each view the lines of sight (LoS) are generated using the intrinsic and extrinsic camera parameters. In \cref{fig:LoS_visualization}) this is shown for a example camera with 4x3 pixels. While the original RayConv operation only enriched the feature vector with the ray direction, our approach also encodes the origin leading to 6 values, 3 for the origin and 3 for the direction of the ray. This leads to a vector of shape $ [\text{batch} , \text{num}\_\text{views}, H_l , W_l, 6] $ where $ H_l $ and $ W_l $ correspond to height and width of feature level $ l $. The feature vector generated by the encoder is concatenated with the encoded lines of sight and is then projected back to the encoding dimension using a linear layer.  We call this operation FLoSE (Feature Line-of-Sight Encoding), which can be defined as follows:
\begin{equation}
    \hat{\mathbf{F}}_{v} = \operatorname{Concat}\left( \mathbf{F}_{v} , \mathbf{R}_{v}  \right) \mathbf{W}^\top \enspace .
\end{equation}
$\mathbf{F}_{v} \in \mathbb{R}^{X, d_\mathrm{model}}$ are the encoded multi-scale features and $ \mathbf{R}_{v} \in \mathbb{R}^{X, 6} $ are the parametrized lines of sight. The $ X $ stands for the dimension of $ [\text{batch} \times \text{num}\_\text{views}, \sum_l H_l \times W_l] $. After concatenation, they are transformed back to the embedding dimension with the learned weights $\mathbf{W}^\top \in \mathbb{R}^{d_\mathrm{model} , (6 + d_\mathrm{model} ) }$. The resulting spatially enriched feature vector $\hat{\mathbf{F}}_{v} \in \mathbb{R}^{X, d_\mathrm{model} }$ can then be used within the projective attention module to sample local features in the vicinity of the objects.

The model needs to sample features of the vector $ \hat{\mathbf{F}}_{v} $. As described in Deformable-DETR \cite{zhuDeformableDETRDeformable2020}, this can be done with so-called reference points. Within the projective attention module of \cite{wangDirectMultiviewMultiperson2021}, these points are generated by projecting 3D joint locations into the corresponding 2D image locations using the camera orientation parameters. We adapted this module to work within the setup of multi-view object pose estimation. The object detector already provides suitable reference points for each view, namely the center values of the bounding boxes of the objects. 
The features are then collected from $K$ discrete sampling points away from the center $ \mathbf{c} $. These sampling offsets are estimated by projecting the query feature $ \mathbf{q} $ and the spatially enriched feature $ \hat{\mathbf{F}}_{v} (\mathbf{c}) $ with the learnable linear weight $ \mathbf{W}^p $, i.e.,
\begin{equation}
    \Delta \mathbf{c} = (\mathbf{q} + \hat{\mathbf{F}}_{v}(\mathbf{c})) \mathbf{W}^p \enspace .
\end{equation}
At these sampling locations, the corresponding $ \hat{\mathbf{F}}_{v} $ are extracted and an attention weight $ \mathbf{a} = \operatorname{Softmax}((\mathbf{q} + \hat{\mathbf{F}}_{v}(\mathbf{c}) \mathbf{W}^a ) $ is applied linearly. This leads to the following definition for the projective attention:

\begin{equation}
    \begin{aligned}
        \operatorname{PAttention}\left(\mathbf{q}, \mathbf{p},\left\{\hat{\mathbf{F}}_{v}, v=1, \ldots, V \right\}\right) \\
        = \operatorname{Concat}\left(\mathbf{f}_{1}, \mathbf{f}_{2}, \ldots, \mathbf{f}_{V}\right) \mathbf{W}^{p} \enspace ,
  \end{aligned}
\end{equation}
where
\begin{equation}
    \mathbf{f}_{v} =\sum_{k=1}^{K} \mathbf{a}(k) \cdot \hat{\mathbf{F}}_{v} \left(\mathbf{c} + \Delta \mathbf{c}(k)\right) \mathbf{W}^{f} \enspace .
\end{equation}
Here, $\mathbf{W}^p$ and $\mathbf{W}^f$ are learnable linear weights. The projective attention is applied within each decoder layer, where the cross-attention can learn relationships between the different views.

The decoder outputs object queries, which are processed by the rotation and translation heads. Both heads are multi-layer perceptrons (MLPs) and remain unchanged with respect to the implementation in PoET \cite{jantosPoETPoseEstimation2023}. This means that the translation head predicts the three translation values $\mathbf{t}^p = (t^p_x, t^p_y, t^p_z)$ for the reference view. Consequently, the loss is determined as in PoET \cite{jantosPoETPoseEstimation2023}:
\begin{equation}
    L_\mathrm{t} = \|\mathbf{t} - \Tilde{\mathbf{t}}\|_2 \enspace .
\end{equation}
The rotation head directly predicts a 6D rotation representation from which the 3$\times$3 rotation matrix can be recovered using the Gram-Schmidt orthogonalization by interpreting the 6 values as two vectors. The 6D representation ensures a stable training with the least number of values. This is described in detail in \cite{zhouContinuityRotationRepresentations2019} where it was explained that such a 6D representation is favorable. The loss is calculated via the trace of the matrix product of prediction $ \Tilde{\mathbf{R}} $ and ground truth $ \mathbf{R} $ rotation:
\begin{equation}
    L_\mathrm{rot} = \arccos\frac{1}{2} \left( \operatorname{Tr} \left( \mathbf{R} \Tilde{\mathbf{R}}^{T} \right) - 1 \right) \enspace .
\end{equation}
Both losses are weighted with $\lambda_\mathrm{t}$ and $\lambda_\mathrm{rot}$ to obtain the total loss of
\begin{equation}
    L = \lambda_\mathrm{rot} L_\mathrm{rot} + \lambda_\mathrm{t} L_\mathrm{t} \enspace .
\end{equation}

In \cite{jantosPoETPoseEstimation2023}, two different approaches with respect to the number of classes are compared. In the class-specific case, the output dimensions of the translation and rotation head are $3 \times n_\mathrm{cls}$ and $6 \times n_\mathrm{cls}$, respectively, while in the class-agnostic case, the dimensions are 3 and 6 for translation and rotation. Because \cite{zhouContinuityRotationRepresentations2019} describes that the class-specific case yielded better results, we continued with this approach.

\section{EXPERIMENTS}
\label{sec:Experiments}
In this section, the experiments we have conducted are explained. Of the existing multi-view methods, only the code of CosyPose \cite{labbeCosyPoseConsistentMultiview2020} is publicly available. Therefore, it is the only multi-view method we could test on the MV-ball dataset. Furthermore, we evaluated PoET \cite{jantosPoETPoseEstimation2023} on the MV-ball dataset since it is a SOTA pose estimation algorithm and served as a basis for our approach. For the YCB-V dataset additional approaches were compared with our approach.

\subsection{Metric}
To evaluate our model, we used the mean ADD-(S) error introduced by \cite{hinterstoisserModelBasedTraining2012} and the rotation/translation error in \cite{billingsSilhoNetRGBMethod2019} for the MV-ball dataset. The rotation error is of particular interest since it can better highlight the benefits of our multi-view approach if applied on the MV-ball dataset. On the YCB-V dataset, we use the AUC of the ADD-S metric since this is a commonly used metric for this dataset.

\subsection{Datasets}
As mentioned in \cref{sec:Dataset}, we developed our own dataset MV-ball to prove the true multi-view nature of our approach. We investigated the influence of the visibility of the hemispheres with the two different test sets.

Furthermore, we evaluated our model on the YCB-V dataset \cite{xiangPoseCNNConvolutionalNeural2018}, which is a well established pose estimation dataset and also part of the BOP Challenge \cite{hodanBOPChallenge20202020,nguyenBOPChallenge20242025} (see \cref{sec:Related works}). Our model requires the interior and relative orientations of the cameras. In the YCB-V dataset, only the real split of the training data provides the external camera parameters. We therefore only trained a single view version of our model but trained on the real and synthetic images to allow a fair comparison.

\subsection{Implementation Details}
Our baseline model has 5 encoder/decoder layers, 16 heads, and an embedding dimension of 256. As the backbone, we use a Mask R-CNN \cite{heMaskRCNN2017} for the MV-ball dataset and YOLOv4 \cite{wangScaledYOLOv4ScalingCross2021} for the YCB-V dataset. We used the object detector backbone only for multi-scale feature extraction and used the ground-truth labels for the bounding box information. The use of the object detector for bounding box prediction is not meaningful when evaluating the pose estimation pipeline itself. In the case of MV-ball, we added a random jitter as introduced by PoET. The lines of sight are encoded as direction + origin. We trained our baseline model for 100 epochs using the AdamW optimizer with a learning rate \num{2e-5} and a batch size of 1. For YCB-V, we trained for 100 epochs at a batch size of 16. In PoET, the number of queries was set based on the maximum number of objects in an image \cite{jantosPoETPoseEstimation2023}. Consequently, we used 2 queries for MV-ball and 10 for YCB-V. However, we investigated different numbers of queries in the ablations. 

\begin{figure}[!t]
    \centering

    \begin{minipage}[b]{0.48\linewidth}
        \centering
        \includegraphics[width=\linewidth]{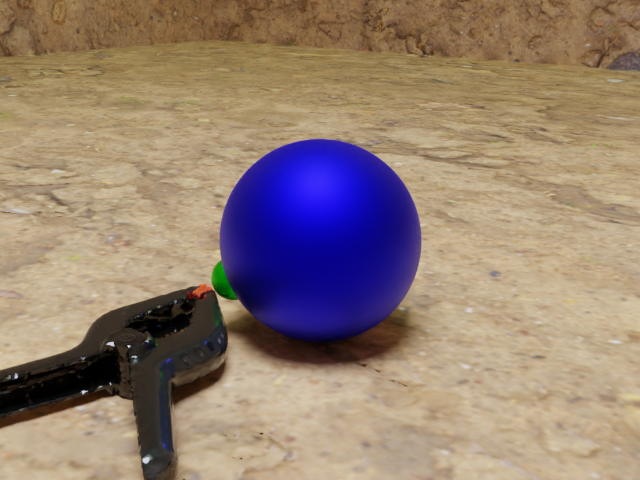}
        \subcaption{View 1}  
        \label{fig:view1_blank}
    \end{minipage}
    \hfill
    \begin{minipage}[b]{0.48\linewidth}
        \centering
        \includegraphics[width=\linewidth]{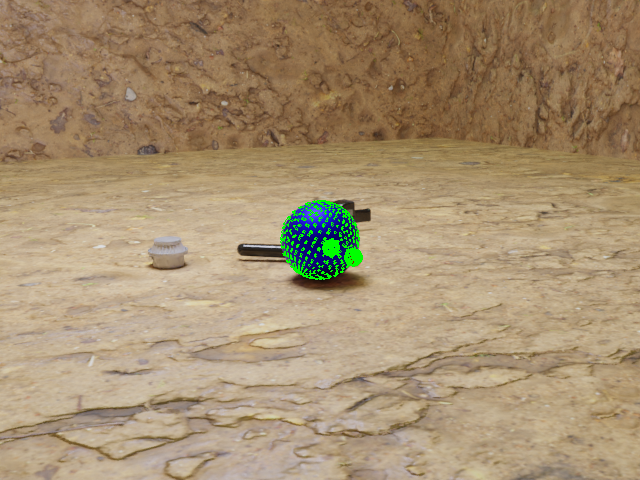}
        \subcaption{View 2 + gt overlay (green)}  
        \label{fig:view2_gt}
    \end{minipage}

    \vspace{0.5em} 

    \begin{minipage}[b]{0.48\linewidth}
        \centering
        \includegraphics[width=\linewidth]{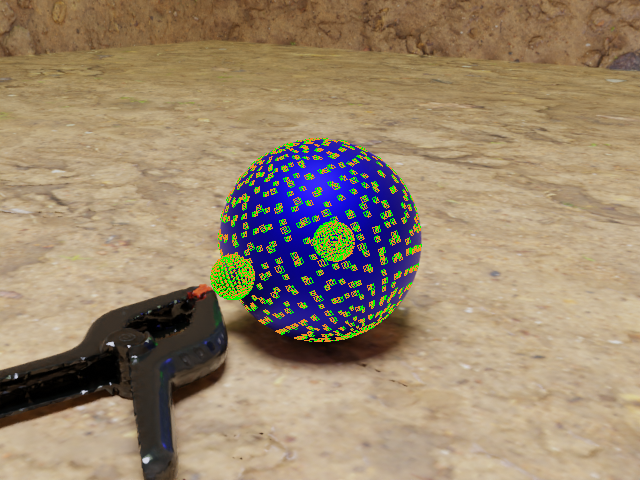}
        \subcaption{View 1 + gt/prediction (Ours)}  
        \label{fig:view1}
    \end{minipage}
    \hfill
    \begin{minipage}[b]{0.48\linewidth}
        \centering
        \includegraphics[width=\linewidth]{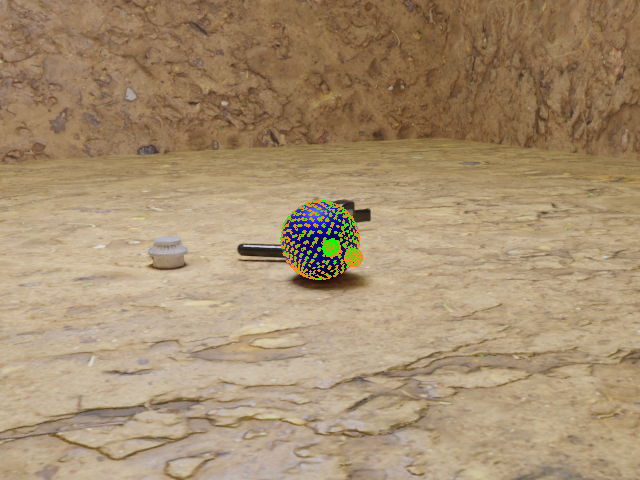}
        \subcaption{View 2 + gt/prediction (Ours)}  
        \label{fig:view2}    
    \end{minipage}

    \vspace{0.5em} 

    \begin{minipage}[b]{0.48\linewidth}
        \centering
        \includegraphics[width=\linewidth]{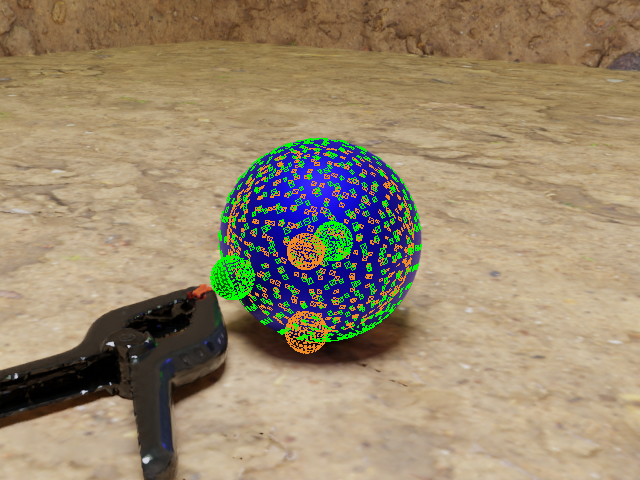}
        \subcaption{View 1 + gt/prediction (Cosypose)}  
        \label{fig:view1_cosy}
    \end{minipage}
    \hfill
    \begin{minipage}[b]{0.48\linewidth}
        \centering
        \includegraphics[width=\linewidth]{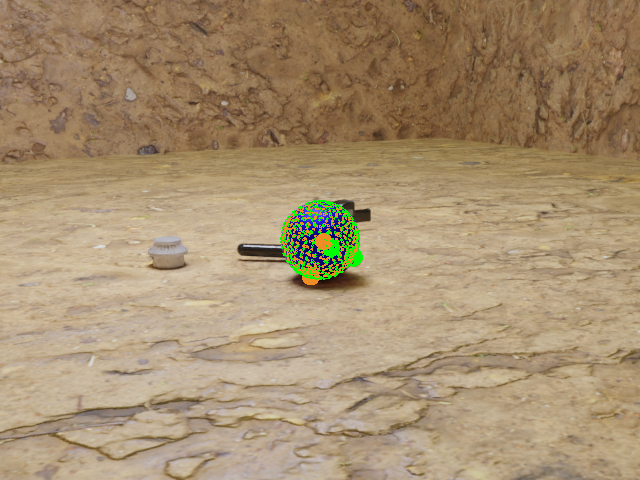}
        \subcaption{View 2 + gt/prediction (Cosypose)}  
        \label{fig:view2_cosy}    
    \end{minipage}

    \caption{Example view pair of the easy test set of the MV-ball dataset. a) shows view 1 without any overlays. b) shows view 2 with the ground-truth (gt) overlay (green) only. In c) and d), both views are displayed with prediction of MVTOP (orange) and gt (green) while e) and f) show corresponding results with Cosypose}
    \label{fig:mvball_pred_gt}
\end{figure}

\subsection{Results}

\begin{table*}[t]
    \caption{Results of different methods on the MV-ball dataset. The ADD error is reported in mm and the rotation error in degree. CosyPose and MVTOP (Ours) are trained/evaluated with one view (1v) and two views (2v)}
    \centering
    \begin{tabular}{c|ccc}
    \toprule
    Method & Mean ADD & Mean rot. error \\
    \midrule
    PoET \cite{jantosPoETPoseEstimation2023} & 0.07552 & 95.455 \\ 
    CosyPose 1v \cite{labbeCosyPoseConsistentMultiview2020} & 1.04291 & 105.539 \\ 
    CosyPose 2v \cite{labbeCosyPoseConsistentMultiview2020} & 1.04312 & 105.539 \\ 
    \bottomrule
    Ours 1v & \textbf{0.03702} & \textbf{41.362 } \\ 
    Ours 2v & \textbf{0.01185} & \textbf{7.345} \\ 
    \bottomrule
    \end{tabular}
    \label{tab:res_mv_ball}
\end{table*}

The results of the MV-ball dataset can be seen in \cref{tab:res_mv_ball}. We achieve a mean ADD error of 0.01185\thinspace m and a mean rotation error of $ 7.345 \degree$. Therefore, we outperform PoET \cite{jantosPoETPoseEstimation2023} and CosyPose \cite{labbeCosyPoseConsistentMultiview2020} by a large margin. Both methods do not resolve the pose ambiguity described in \cref{sec:Dataset}. For PoET, this is obvious because it is a single-view method and these methods fail by design. For CosyPose, the situation is similar since the multi-view aspect in CosyPose is based on the poses estimated from the single views. The approach of \cite{liUnifiedFrameworkMultiView2018} cannot be evaluated on this dataset because the code is unavailable.

The qualitative results of our approach can be seen in \cref{fig:mvball_pred_gt}. Here, the 3D model was used to generate a mesh that was projected into the 2D image using the interior orientation of the camera and the predicted/gt pose. For the sake of visibility, only every 8th mesh vertex was projected, resulting in an appearance that resembles a point cloud. In \cref{fig:view1} and \cref{fig:view2}, the projected prediction (orange) and the ground-truth (gt) projection (green) align so well that they barely can be distinguished. This highlights the performance of our multi-view approach. It is important to note that due to the projection of the complete 3D model into the image plane, the mesh of the hidden hemisphere is displayed although the hemisphere is not visible in the image itself. In \cref{fig:view1_blank}, it can be seen that only the green hemisphere (close to the black clamp) is visible. In \cref{fig:view1}, the projection of the mesh shows that the red hemisphere (denser green/orange mesh points in the middle) is located on the other side of the ball. Conversely, in \cref{fig:view2}, the red hemisphere would be visible (although it cannot be seen due to the overlay of gt and prediction), while the location of the green hemisphere is indicated again by the denser green/orange mesh points almost in the middle of the ball. The corresponding qualitative results of Cosypose can be seen beneath.

\begin{figure}[!t] 
    \begin{minipage}[b]{0.98\linewidth}
        \centering
        \includegraphics[width=1\linewidth]{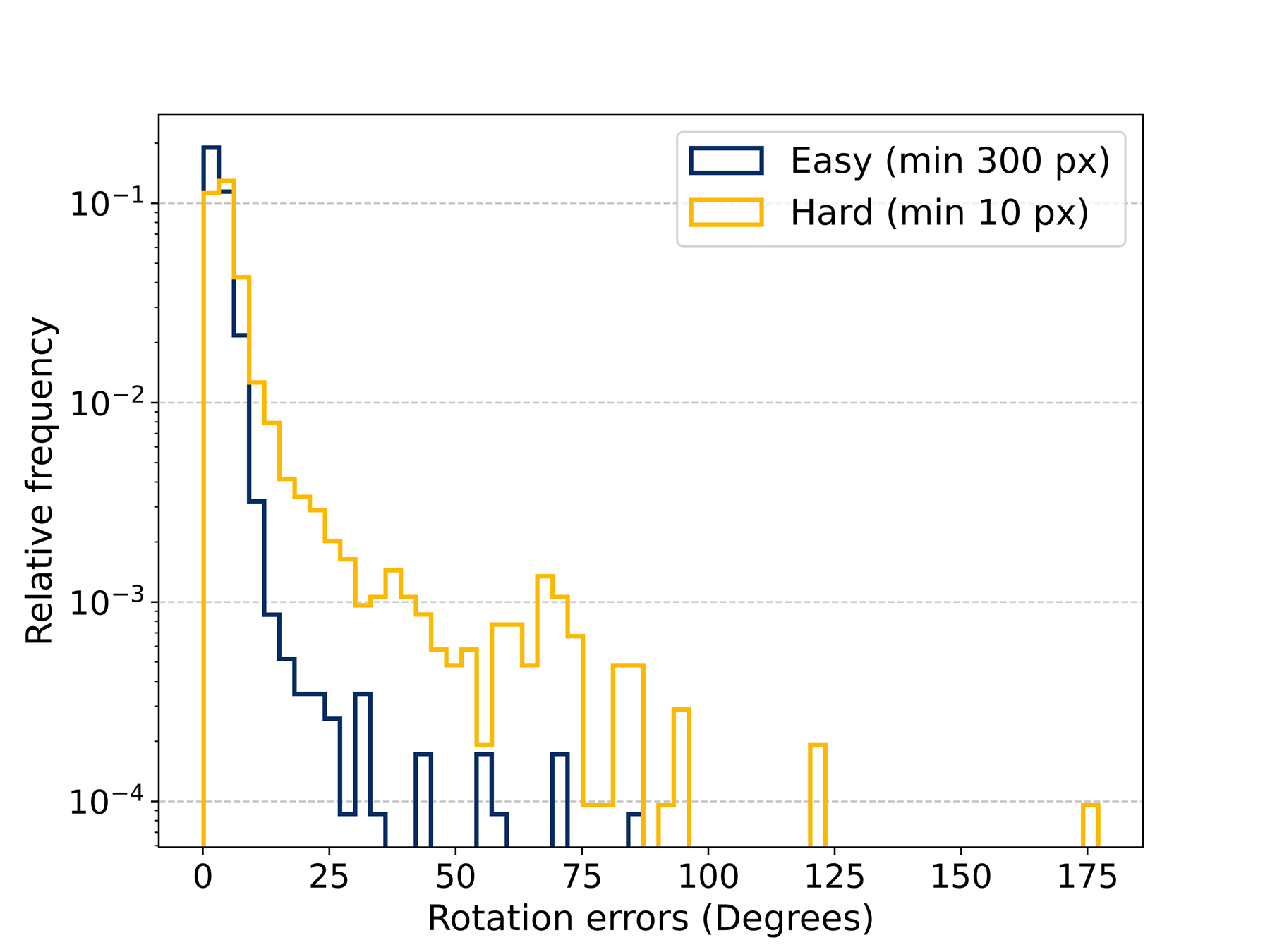}
    \end{minipage}
    \caption{Histogram for the rotation errors of MVTOP on the two different test sets of the MV-ball dataset. For the easy set (blue) and the hard set (yellow), a mean rotation of 3.43° and 7.39° was reached. Note the logarithmic scale of the $y$ axis.}
    \label{fig:rotation_plot}
\end{figure}
As described in \cref{sec:Dataset}, we generated two test sets of the MV-ball to investigate the influence of the visibility of the hemispheres. The results can be seen in the histogram of \cref{fig:rotation_plot}. The bin size corresponds to a step size of 3°. The easy set consists of 3853 images and the hard set of 3458 images. It can be seen that a higher threshold for the visibility of the hemispheres clearly reduces the rotation error, leading to a better pose estimation. For the easy test set with a threshold of 300 pixels, the highest rotation error is at $\approx 85^\circ$ and the mean is $\approx 3.43^\circ$ while for the hard test set with a threshold of 10 pixels the highest error is $\approx 174^\circ$ and the mean is $\approx 7.39^\circ$.

The results of the YCB-V dataset can be seen in \cref{tab:res_YCBV}. We compare our method with PoET \cite{jantosPoETPoseEstimation2023}, T6D \cite{aminiT6DDirectTransformersMultiobject2021}, GDR-Net \cite{wangGDRNetGeometryGuidedDirect2021}, ZebraPose \cite{suZebraPose2022IEEE2022}, CosyPose \cite{labbeCosyPoseConsistentMultiview2020} and the appraoch of \cite{liUnifiedFrameworkMultiView2018}. We achieve SOTA results with an AUC of the ADD-S metric of 96.50. However, as discussed in \cref{sec:YCB-V_flaw}, the synthetic training split suffers a flaw by incorporating test information. Therefore, results of methods that are trained on the synthetic YCB-V split cannot be compared in a fair manner and might even be meaningless. In this comparison, this would apply to the methods of \cite{jantosPoETPoseEstimation2023,aminiT6DDirectTransformersMultiobject2021,liUnifiedFrameworkMultiView2018} and MVTOP. 

\begin{table}[h]
    \caption{Results of different methods on the YCB-V dataset using the AUC of ADD-S metric.}
    \centering
    \begin{tabular}{c|ccc}
    \toprule
    Method & AUC of&ADD-S & ADD-(S)  \\
    \midrule
    PoET & & 92.8 & - \\
    T6D & & 86.2 & 74.6 \\
    GDR-Net & & 91.6 & 84.4 \\
    ZebraPose & & 90.1 & 85.3 \\
    Li et al., 1v & & 75.1 & - \\
    Li et al., 5v & & 80.2 & - \\
    CosyPose 1v & & 89.8 & 84.5 \\
    CosyPose 5v & & 93.4 & - \\
    \bottomrule
    Ours & & \textbf{96.50} & \textbf{92.16} \\
    \bottomrule
    \end{tabular}
    \label{tab:res_YCBV}
\end{table}

\subsection{Runtime Analysis}

\begin{table}[h]
    \caption{Runtime analysis (in ms) for a Nvidia GTX 1080 (GPU 1) and a Nvidia RTX 3080 (GPU 2). The analysis was conducted for 1,2 and 4 views.}
    \centering
    \begin{tabular}{ccc|cc}
    \toprule
     \# Views & GPU 1 & GPU 2 & CosyPose & T6D \\
    \midrule
    1 & 82.90 & 69.97 & & 170 \\
    2 & 131.51 & 100.96 & &  \\
    4 & 224.66 & 163.39 & 320 & \\
    \bottomrule
    \end{tabular}
    \label{tab:res_runtime}
\end{table}

\noindent
The runtime analysis was performed on the YCB-V dataset with a model only trained on the real-data training split since we could test different numbers of views. We measured the inference time per image. To reduce the influence of processes like initialization, data loading, etc.\ on the mean value, we averaged five runs per test. As the test set, we used the subset of the YCB-V testset proposed by the BOP-Challenge \cite{hodanBOPChallenge20202020}, which consists of 75 images from each of the 12 test scenes totaling 900 validation images at a resolution of 640$\times$480 pixels with a total of 4125 annotations. Therefore, each image contained on average $\approx4.6$ objects. For comparison, we cite the runtime of CosyPose and T6D as described in the papers \cite{labbeCosyPoseConsistentMultiview2020,aminiT6DDirectTransformersMultiobject2021}. It is important to note that a complete and fair comparison cannot be made because it is not stated in \cite{labbeCosyPoseConsistentMultiview2020,aminiT6DDirectTransformersMultiobject2021} which hardware was used and on which dataset or which image size the analysis was performed. 

However, our tests indicate that our model runs at a competitive speed.
Doubling the number of views results in a increased inference time by a factor of 1.59/1.71 for the GTX 1080 and 1.44/1.62 for the RTX 3080. Thus, doubling the number of input images leads to an average inference increase by a factor of 1.59 which shows that model does not scale linearly with the number of views.

\section{ABLATIONS}
\label{sec:Ablations}
The ablations were performed on the MV-ball dataset. While in PoET \cite{jantosPoETPoseEstimation2023} a classical encoder-decoder approach was used, \cite{wangDirectMultiviewMultiperson2021} only used a decoder that operates directly on the feature maps. Therefore, we investigated the influence of the encoder and present the results in \cref{tab:abl_enc}. It can be seen that without the encoder, the mean ADD and rotation error increase. This shows the strong contribution of the encoder in generating meaningful embeddings of view-specific features.

\begin{table}[h]
    \caption{Effect of the encoder in the MVTOP network}
    \centering
    \begin{tabular}{cccc}
    \toprule
     encoder & mean ADD & mean Rotation Error \\
    \midrule
    w/ & 0.01185 & 7.345 \\ 
    w/o & 0.03287 & 33.667 \\ 
    \bottomrule
    \end{tabular}
    \label{tab:abl_enc}
\end{table}
As described in \cref{sec:Method}, the model is enriched with the embedding of the lines of sight (LoS) of the different views. However, there are several ways to encode the LoS. Our baseline encoded them as the origin and direction of the vectors for the pixels. Our ablations include three additional encoding approaches. That is, just giving the direction of the rays, encoding the rays via the Plücker coordinates, and passing the Plücker coordinates and the origin. The Plücker coordinates consist of the direction of the rays and the so-called moment, which is calculated as the vector product of the direction of the ray $ \vec{\mathbf{d}}  $ and one point lying on the ray and by using the origin $ \vec{\mathbf{o}} $ as this point the moment can be calculated as follows \cite{dorstGeometricAlgebraComputer2007}:
\begin{equation}
    \mathbf{\mathbf{m}} = \vec{\mathbf{d}} \times \vec{\mathbf{o}}
\end{equation}

\begin{table}[h]
    \caption{Influence of different LoS encodings. From top to bottom: direction + origin, direction only, Plücker coordinates and Plücker coordinates + origin}
    \centering
    \begin{tabular}{cccc}
    \toprule
     LoS encoding & Mean ADD & Mean rot. error \\
    \midrule
    baseline & 0.01185 & 7.345 \\ 
    dir only & 0.01387 & 10.267 \\ 
    Pl. coord. & 0.014 & 10.858 \\ 
    Pl. coord. + orig & \textbf{0.01042} & \textbf{6.211} \\ 
    \bottomrule
    \end{tabular}
    \label{tab:abl_LoS_encoding}
\end{table}
\Cref{tab:abl_LoS_encoding} shows the results of the different LoS encodings. It can be seen that the mean ADD and rotation error do not vary much with respect to the baseline. While passing the origin alone or Plücker coordinates worsened the errors, the performance of our model could be increased somewhat by passing Plücker coordinates plus the origin. We assume that, although the origin is encoded in the moment of the Plücker coordinates, the model benefits from the direct access to the origin and does not have to learn to unravel the vector product.

A key component of the transformer are the queries for which the model makes its prediction. Consequently, the number of queries defines the upper bound of the number of objects to detect. In the original Deformable-DETR setup \cite{zhuDeformableDETRDeformable2020}, the number of queries was set to 300, which led to an accuracy on the COCO dataset of 43.2\%. In \cite{zhangDenseDistinctQuery2023}, the queries were increased, to 900 which led to an accuracy of 45.4\%. This indicates that more queries can lead to better performance. In PoET, the number of queries was set as the maximum number of objects that can appear in an image plus one additional query. Because the YCB-V dataset has at most nine objects in an image, the number of queries in PoET was set to ten. We wanted to investigate whether the number of queries plays a crucial role with respect to accuracy. Corresponding to PoET, we started from our baseline, which used two queries since the MV-ball dataset always contained one ball per image. We examined four other implementations, namely with one query, four queries, and eight queries. The results can be seen in \cref{tab:num_queries}.

\begin{table}[h]
    \caption{Influence of the number of queries}
    \centering
    \begin{tabular}{cccc}
    \toprule
     \# queries & Mean ADD & Mean rot. error \\
    \midrule
    baseline (2) & 0.01185 & 7.345 \\ 
    1 & 0.01089 & \textbf{6.904} \\ 
    4 & 0.01091 & 7.139 \\ 
    8 & \textbf{0.0095664} & 7.139 \\ 
    \bottomrule
    \end{tabular}
    \label{tab:num_queries}
\end{table}
It can be seen that the errors barely change. Furthermore, the best results differ depending on the metric used. While the relative difference in the number of queries corresponds to the mentioned works above (a factor of 2--4), the absolute change is small. 

\begin{figure*}[!t] 
    \centering
    \begin{minipage}[b]{0.8\linewidth}
        \includegraphics[width=1\linewidth]{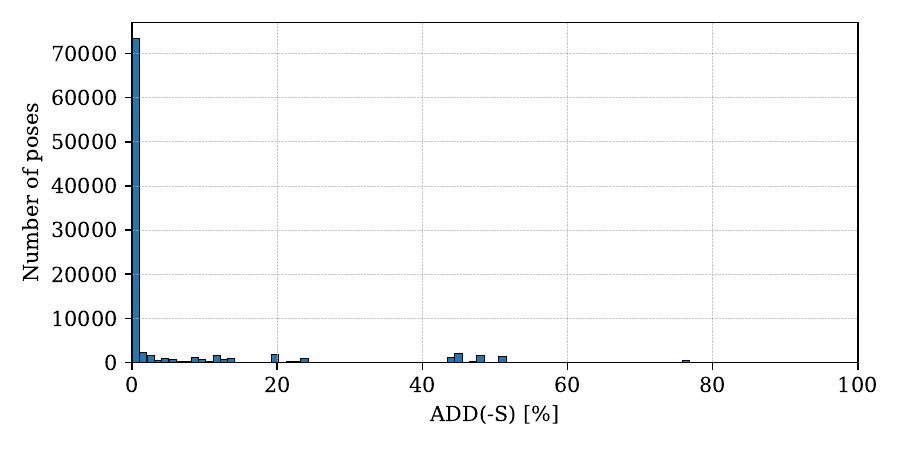}
    \end{minipage}
    \caption{Comparison of all test poses with all poses of the synthetic training split taking the object ID into account. The histogram shows the distribution of test poses with respect to the difference in the ADD-(S) metric to the nearest matching pose. The x-axis has a bin size of $ 1\% $.
  }
  \label{fig:ycbv_histogram}
\end{figure*}

\section{LIMITATIONS and FUTURE WORK}
\label{sec:Limitations}
As described in \cref{sec:Method}, our method defines queries based on the bounding boxes of a reference image. The prediction is made for these queries, i.e., only for the first image. All other images/views provide only additional information, which means that no prediction is made for these images directly. As mentioned in \cref{sec:Method}, we only work with the gt annotations; therefore, no detections are missed. With a detector, this would not reflect reality due to, e.g., occlusions. This could lead to the situation where the reference image only has three detections while an additional image/view has six detections. This could be fixed by generating the queries/input based on the number of detections of the images. Furthermore, it would be possible that a camera setup generates several non-overlapping areas. Therefore, it would be possible to have, e.g., two objects in two different non-overlapping regions. Both views could only detect one object and a generation of the queries based on the number of detections would not lead to more possible true predictions. In this case, the queries would need to be generated for every view, so predictions are made for every view separately. On one hand, this would ensure that objects that fall into such non-overlapping regions can also be found, while on the other hand, the computational cost would scale even more with the number of views. This can be useful depending on the setup and task. In short, in these cases, our method would always miss detections of certain areas, depending on the reference view.

Future work could not only tackle the aforementioned limitations but also extend this or other methods following that architecture to be more applicable in real world scenarios. It can be considered that in a setup, cameras with different resolution, aspect ratios, focal lengths etc. are used. Furthermore, it is desirable to omit the necessity to pretrain the model not only on the objects of interest but also on the number of used cameras/views. Additionally, the latest improvements in DETR-based approaches could be distilled to gain both speed and accuracy.

\begin{figure*}[!t]
    \centering

    \begin{minipage}[b]{0.40\linewidth}
        \centering
        \includegraphics[width=\linewidth]{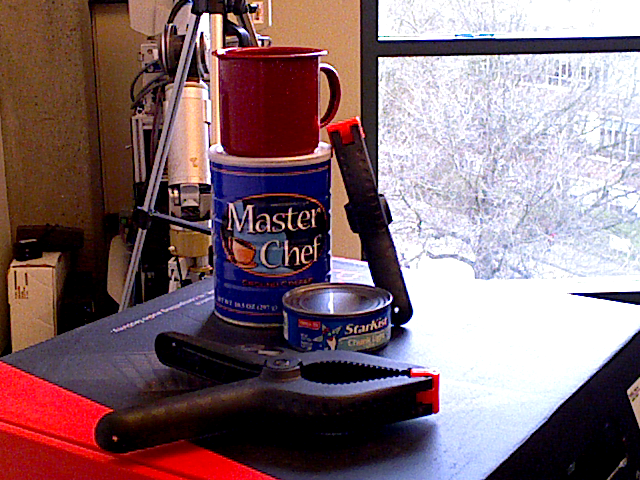}
        \subcaption{Image 000048/000002 of the YCB-V test split}  
    \end{minipage}
    \hspace{0.1\textwidth}
    \begin{minipage}[b]{0.4\linewidth}
        \centering
        \includegraphics[width=\linewidth]{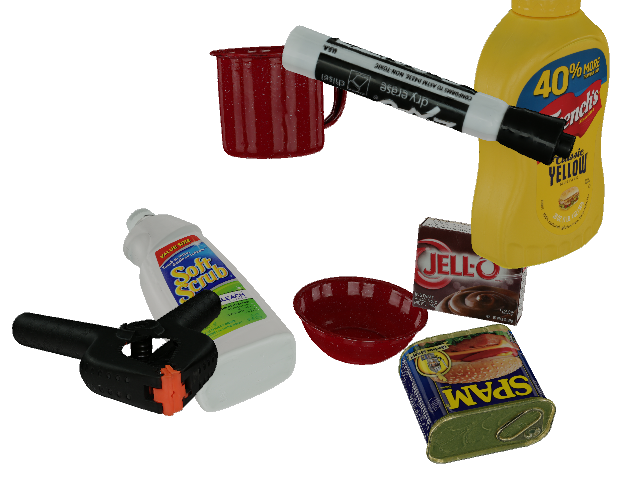}
        \subcaption{Image 000002/002468 of the YCB-V synthetic split)}  
    \end{minipage}

  \caption{Qualitative comparison of the test split and the synthetic split of the YCB-V dataset. The poses of the red cup are practically identical.}
  \label{fig:YCB_V_flaw}
\end{figure*}

\section{YCB-V TRAINING DATA FLAW}
\label{sec:YCB-V_flaw}

During our experiments we found a potentially critical flaw in the YCB-V training dataset ("train\_synth"). A significant amount of the object poses in the synthetic training images were taken from the test set. The objects are not cut out from the test set, but are rendered from the 3D-model, which results in a similar appearance as can be seen in \cref{fig:YCB_V_flaw}. The translation and rotation of the cup in the image from test split differ from the cup in the image from the synthetic split in \cref{fig:YCB_V_flaw} by \num{5.29623e-4} mm and \num{7.426e-5} degrees. This indeed indicates that the exact poses from the test set were taken to generate poses for the synthetic set.

To highlight our observations, we numerically investigated the individual poses. We compared all translation vectors of an object ID in the test set with all vectors of the same object ID in the synthetic dataset. With a maximum distance of both vectors of $ 0.01 $\thinspace mm, we found that $ \approx 30\% $ of the poses in the test set have a duplicate in the synthetic set which corresponds to $\approx 8\% $ of the poses of the synthetic set being taken as poses from the test split. The same applies for the real training split of the YCB-V dataset. We found that $\approx 32\% $ of the poses in the training set have a duplicate in the synthetic dataset, which corresponds to $ \approx 59\% $ of the poses of the synthetic set being taken as poses from the real split. In total, $ \approx 67\% $ of the synthetic poses are either taken from the test set or the real set with respect to the translation error and the mentioned threshold. We used this threshold since it corresponds to float precision with respect to the typical size of the translations which lie in the range of 1000\thinspace mm. We also examined the ADD difference of the poses found with the aforementioned translation threshold. In \cref{fig:ycbv_histogram}, the histogram of the differences in the ADD metric of the poses is displayed with a bin size of 1\%. Over 70,000 poses from the test set have a duplicate pose in the synthetic training set with an ADD difference of at most 1\%. With 98,567 poses in total, this corresponds to over 71\%. Therefore, the synthetic training set is heavily corrupted by information from the test set. A method that overfits and only produces poses from the training set could still find over 70\% of all test poses correctly.

To the best of our knowledge, this has not been acknowledged since the dataset was introduced eight years ago. This means that methods that were also trained on this synthetic data split are not trained correctly and cannot be compared in a fair manner with each other and especially not with methods that do not use this data at all. This renders the corresponding results on this dataset meaningless.

\section{CONCLUSION}
\label{sec:Conclusion}
In this work, we presented a novel multi-view pose estimation network. The attention-based encoder--decoder approach fuses multi-view features and enriches the embeddings with line-of-sight information. This enables the model to establish a strong understanding of the scene.

Furthermore, we created the MV-ball dataset that allows us to benchmark pose estimation methods with respect to their multi-view capability. The poses of the instances in the dataset can only be determined if the view-specific information is combined before the actual pose estimation.

Our method MVTOP can resolve the mentioned pose ambiguity and outperforms existing pose estimation methods on this task. As far as we know, no other existing method can resolve such pose ambiguities consistently. Furthermore, our approach achieves SOTA results on the well-established YCB-V dataset. However, the evaluation results on the YCB-V dataset might only be of limited use because of the flaw in the dataset we pointed out in this work as well.

\newpage
\bibliographystyle{apalike}
{\small
\bibliography{references_v2_bibtexstyle_better}}
\end{document}